\documentclass[conference]{IEEEtran}
\usepackage{graphicx}
\usepackage{amsmath}
\usepackage{booktabs}
\usepackage{tabularx}
\usepackage{hyperref}
\usepackage{caption}
\usepackage{subcaption}
\usepackage{array}
\usepackage{multirow}
\usepackage{float}

\begin{document}

\title{Misinformation Detection using Large Language Models with Explainability}

\author{
\IEEEauthorblockN{Jainee Patel\IEEEauthorrefmark{1}, Chintan M. Bhatt\IEEEauthorrefmark{2}, Himani Trivedi\IEEEauthorrefmark{1}, Thanh Thi Nguyen\IEEEauthorrefmark{3}}
\IEEEauthorblockA{\IEEEauthorrefmark{1}Department of Computer Engineering, LDRP Institute of Technology and Research,\\
Kadi Sarva Vishwavidyalaya, Gandhinagar, India\\
Email: jaineepatel2426@gmail.com, himani\_ce@ldrp.ac.in}
\IEEEauthorblockA{\IEEEauthorrefmark{2}University of Wollongong, GIFT City Campus, Gandhinagar, India\\
Email: cbhatt@uow.edu.au}
\IEEEauthorblockA{\IEEEauthorrefmark{3}AiLECS Lab, Faculty of Information Technology, Monash University, Australia\\
Email: thanh.nguyen9@monash.edu}
}
\maketitle

\begin{abstract}
The rapid spread of misinformation on online platforms undermines trust among individuals and hinders informed decision making. This paper shows an explainable and computationally efficient pipeline to detect misinformation using transformer-based pretrained language models (PLMs). We optimize both RoBERTa and DistilBERT using a two-step strategy: first, we freeze the backbone and train only the classification head; then, we progressively unfreeze the backbone layers while applying layer-wise learning rate decay. On two real-world benchmark datasets, COVID Fake News and FakeNewsNet GossipCop, we test the proposed approach with a unified protocol of preprocessing and stratified splits. To ensure transparency, we integrate the Local Interpretable Model-Agnostic Explanations (LIME) at the token level to present token-level rationales and SHapley Additive exPlanations (SHAP) at the global feature attribution level. It demonstrates that DistilBERT achieves accuracy comparable to RoBERTa while requiring significantly less computational resources. This work makes two key contributions: (1) it quantitatively shows that a lightweight PLM can maintain task performance while substantially reducing computational cost, and (2) it presents an explainable pipeline that retrieves faithful local and global justifications without compromising performance. The results suggest that PLMs combined with principled fine-tuning and interpretability can be an effective framework for scalable, trustworthy misinformation detection.
\end{abstract}


\section{Introduction}

The rapid diffusion of misinformation across digital platforms undermines public trust, damages public health, and destabilizes democratic institutions \cite{sonnac2025rebuilding}. Although fact-checking initiatives exist, their manual nature limits scalability, while traditional machine learning (ML) approaches such as logistic regression, Support Vector Machine (SVM), and decision trees struggle to capture the nuanced semantics of deceptive text \cite{ahmad2020fake}. Deep learning methods \cite{zhang2021understanding}, including Convolutional Neural Networks (CNNs) and Recurrent Neural Networks (RNNs), improved representational power but often fail to generalize across domains and languages due to limited context modeling \cite{ali2024comprehensive}.

Conventional detection methods, depending on surface linguistic cues or rule-based heuristics, fail to cope with the dynamic and domain-sensitive characteristics of misinformation. These methods do not have the semantic richness and adaptability required to generalize across domains \cite{guo2019future}. With recent developments in transformer-based large language models (LLMs), detection can potentially be improved by learning deeper contextual relationships in text. These models can be leveraged in both fine-tuning and zero-shot scenarios to detect misinformation with greater adaptability and contextual awareness \cite{kareem2023fighting}.

Moreover, the challenge of misinformation lies in its ability to exploit human psychology and social trust. People are more likely to believe and share information that aligns with their existing views or evokes strong emotions, even without verifying its accuracy. This tendency enables falsehoods to circulate widely before corrections can catch up. As a result, tackling misinformation requires not only technological solutions but also fostering critical thinking, promoting transparency in media, and encouraging responsibility in information sharing~\cite{pennycook2019fighting}.

The proliferation of misinformation in the modern digital era has called into question certain apprehensions about its impact on the opinion, health and stability of democracy. Unlike factual news, rooted on verifiable facts and press skills, misinformation is usually constructed to deceive, performing formulas that disguise bad data as reliable data \cite{swire2020public}. This has especially become evident over the course of global crises such as the COVID-19 pandemic and presidential elections, where misinformation has led to vaccine rejection, mass disinformation efforts, and political unrest \cite{caceres2022impact}. The distinctive feature of misinformation that distinguishes it against mere mistakes is the matter of intention and the design of the content because it is fabricated as to look like an original one when it promotes something false. This type of content is not only able to undermine any public trust with the media but can also destroy the pool of common facts that are needed in order to make good decisions.

Transformer-based pretrained language models (PLMs) like BERT, RoBERTa~\cite{le2024australian} and DistilBERT have made huge leaps in natural language understanding activities, such as fake news and rumor detection \cite{chen2025identifying}. However, several open challenges remain. First, most studies emphasize accuracy while neglecting computational efficiency, an essential factor for real-time deployment in resource-constrained environments. Second, the black-box nature of PLMs undermines user trust, highlighting the need for integrated interpretability. Finally, systematic quantification of efficiency-accuracy trade-offs across lightweight and heavy PLMs for misinformation detection has not been rigorously explored.

\section{Contribution and Novelty}
This study advances misinformation detection by emphasizing \emph{efficiency, transparency, and deployability} rather than purely maximizing accuracy with larger models. Our main contributions are:

\begin{itemize}
    \item \textbf{Lightweight yet competitive modeling.} We demonstrate that DistilBERT, a compact PLM, attains accuracy comparable to RoBERTa on the COVID Fake News dataset while substantially reducing compute. Concretely, our measured DistilBERT training time is $\sim$397\,s/epoch (from training logs) with inference throughput $\sim$71.8 samples/s and latency $\sim$13.9\,ms/sample on 2{,}041 test items. This quantifies a practical path to real-time and edge deployment.
    \item \textbf{Two-phase fine-tuning with Layer-Wise Learning Rate Decay (LLRD).} We adopt a training curriculum that first freezes the backbone to stabilize task adaptation, then progressively unfreezes layers with layer-wise learning rate decay, mitigating catastrophic forgetting and improving convergence.
    \item \textbf{Built-in explainability.} We integrate LIME (local token-level rationales) and SHAP (global attributions), yielding faithful, human-interpretable evidence supporting model decisions, a key requirement for responsible AI in high-stakes applications.
    \item \textbf{Comprehensive evaluation.} Beyond accuracy, we report Precision, Recall, F1, AUROC, and efficiency metrics (parameters, training time/epoch, inference latency, throughput), enabling a holistic comparison to stronger baselines.
\end{itemize}

\section{Related Work}
The fast development of LLMs has introduced a substantial change in the creation and detection of misinformation. Recent studies also suggest that they are useful models, which may be used to identify deceitful content with a significantly higher fidelity than with the use of traditional machine learning or the use of rules set up manually. With the help of their wide pretrained knowledge and logic skills, these models can detect fake news, identifying fake profiles, and crosscheck claims and previous information with very little human oversight \cite{papageorgiou2024survey}. Nevertheless, such a dual capacity is problematic because the same generative powers that result in their being effective detectors can be used to generate convincing disinformation on an unprecedented scale~\cite{augenstein2024factuality}.

In order to mitigate these dangers, more recent research has been dedicated to making better prompt design, particularly using zero-shot and few-shot approaches that nudge models to become more critical in thinking about the veracity of information without the need to undergo massive further training. As an illustration, the application of negative reasoning prompts has revealed that an increase in the size of a model and improvement in instructions may lead to a considerable improvement in the factual accuracy of zero-shot situations. These encouraging outcomes, however, do not mean that there are not limitations when it comes to the extent to which models can generalize across subjects and models can be used to adapt to new situations, which underscores the importance of continuing research to produce effective prompt frameworks and improve upon evaluation strategies~\cite{zhang2025llms}.

The other key advancement in this area is the introduction of multi-hop verification datasets and explainable verification pipelines. Such resources enable models to reduce complicated arguments into small factual processes, secure the components separately, and produce transparent accounts of the verification process. The results indicate that big language models perform better in the majority of conditions when they are used as planners and the generators of justification instead of mere fact categorizers~\cite{ma2023ex}. By focusing on explainability, this area alleviates a need that exists in trust and transparency, where end users and human fact-checkers would have clear opportunities to understand how an output was generated~\cite{ma2023ex}.

Taking into account the global proliferation of fake claim, the multilingual adaptation has also become another priority in research. Multiple-language, large-benchmarking datasets of tens of thousands of claims have been developed to test the extent to which these models can be applied to fact-checking and detection of hallucination in a range of linguistic and cultural settings. Research findings show that the degree of detection can differ broadly according to the constitution of a language, popularity of a subject, as well as presence of information on the internet. Such findings attest to the significance of creating specific solutions to each region that considers local circumstances depending on their deployment in the world~\cite{zhang2025poly}.

Researchers have also shifted their focus to the storage, recall and preservation of factuality within big language models. The relative performance of various model architectures, and in particular retrieval-augmented model configurations, has charted the trade-off between general-purpose scale and domain-specific accuracy. Through reining in standards of factuality and tailored measures, researchers intend to enhance the concept of regularity according to which these paradigms conserve and regurgitate substantively verifiable knowledge with topical normalcy \cite{wang2023survey}. Besides that, there is a budding effort to extend misinformation detection to the multimedia modality, including detecting the falseness in files or disguising style corrections in the subtlest ways~\cite{wu2024fake}. Taken together, these developments point to a future where computers not only possess enhanced abilities to detect misinformation, but also provide transparent, multilingual, and context-aware explanations, strengthening our defense against digital falsehoods.

\section{Methodology}
The misinformation detection pipeline was proposed using a systematic multi-phase approach that allowed achieving high predictability and transparency of the created model.

\subsection{Data Collection and Preprocessing}
Collection and preparation of credible data was the initial step of the process. The COVID Fake News dataset \cite{patwa2021fighting} and the FakeNewsNet GossipCop dataset \cite{jung2023topological} were chosen since these datasets include various, real-world examples of both claiming true and fake news. Systematic cleaning of all the headlines and short text statements was carried out to eliminate unwanted noise, like, hyperlink, special characters, emojis, and HTML tag blocks. The cleaned text is further normalized so that all characters are put into lowercase in order to make the text consistent and enhance the efficiency of the tokenization. This preprocessing procedure was necessary to make sure that the input data had a standardized form that can be used to train modern transformer-based language models \cite{zhang2023large}.

\subsection{Model Selection}
The pipeline tested several state-of-the-art transformer models to be able to capture the fine-grained linguistic patterns~\cite{xue2025improve}. Among them, there are DistilBERT, RoBERTa, the version with social media language called BERTweet-base, and DeBERTa-v3-small. The rich choice has enabled the relativistic comparison to determine the model architecture that brings the most favorable trade-off (accuracy, computational efficiency, explainability).

\subsection{Two-Phase Training Strategy}

We experiment with two transformer PLMs: RoBERTa (high-capacity) and DistilBERT (lightweight). Both are equipped with a task-specific classification head:
\begin{equation}
\hat{y} = \mathrm{Softmax}(W_2 \cdot \mathrm{Dropout}(\sigma(W_1 \cdot h_{[CLS]})))
\end{equation}
where $h_{[CLS]}$ is the pooled embedding of the input sequence, $W_1, W_2$ are trainable weight matrices, and $\sigma$ is the ReLU activation.

The method of training was the best solution to achieve maximum performance without the likelihood of overfitting.

\subsubsection{Phase 1: Feature Extraction}
At first, the backbones of the chosen pre-trained transformer models were frozen. This enabled the high generality of language representations that had been acquired during the pre-training to be maintained. In this stage, only a custom classification head, i.e., an attention mechanism followed by dropout layers, dense layers and a softmax output layer, was trained. This assisted the model in generalizing the process specific to the task of classifying binary misinformation without interfering with the core language comprehension \cite{li2024llm}.

\subsubsection{Phase 2: Fine-Tuning with Layer-wise Learning Rate Decay (LLRD)} 
The second step involved progressively unfreezing all transformer layers. It implements Layer-wise Learning Rate Decay, where lower layers closer to the input embeddings are fine-tuned with smaller learning rates, while higher layers near the classification head are trained with relatively larger learning rates. This incremental unfreezing and variable learning speed mechanism enabled the model to adapt partially to the new training data and resisted catastrophic forgetting of knowledge obtained from older training data~\cite{bu2024layer}.

\subsection{Integration of Explainability}
One of the central targets of this study was related to the fact that the predictions generated by the model should be interpretable by end-users, i.e., fact-checkers and scholars. In the case of highlighting local interpretability, the Local Interpretable Model-Agnostic Explanations (LIME) approach had been used to gain insights into what precise tokens or phrases in an input text were influential within the final prediction. On the other hand, the SHapley Additive exPlanations (SHAP) method was chosen to calculate and plot (global interpretability) the sums of the input of various features in the whole data. All these explainability methods cumulatively instill certainty that the outputs of the system are clear, comprehensible, and reliable~\cite{xu2019explainable}.

\begin{figure}[ht]
    \centering
    \includegraphics[width=0.6\linewidth]{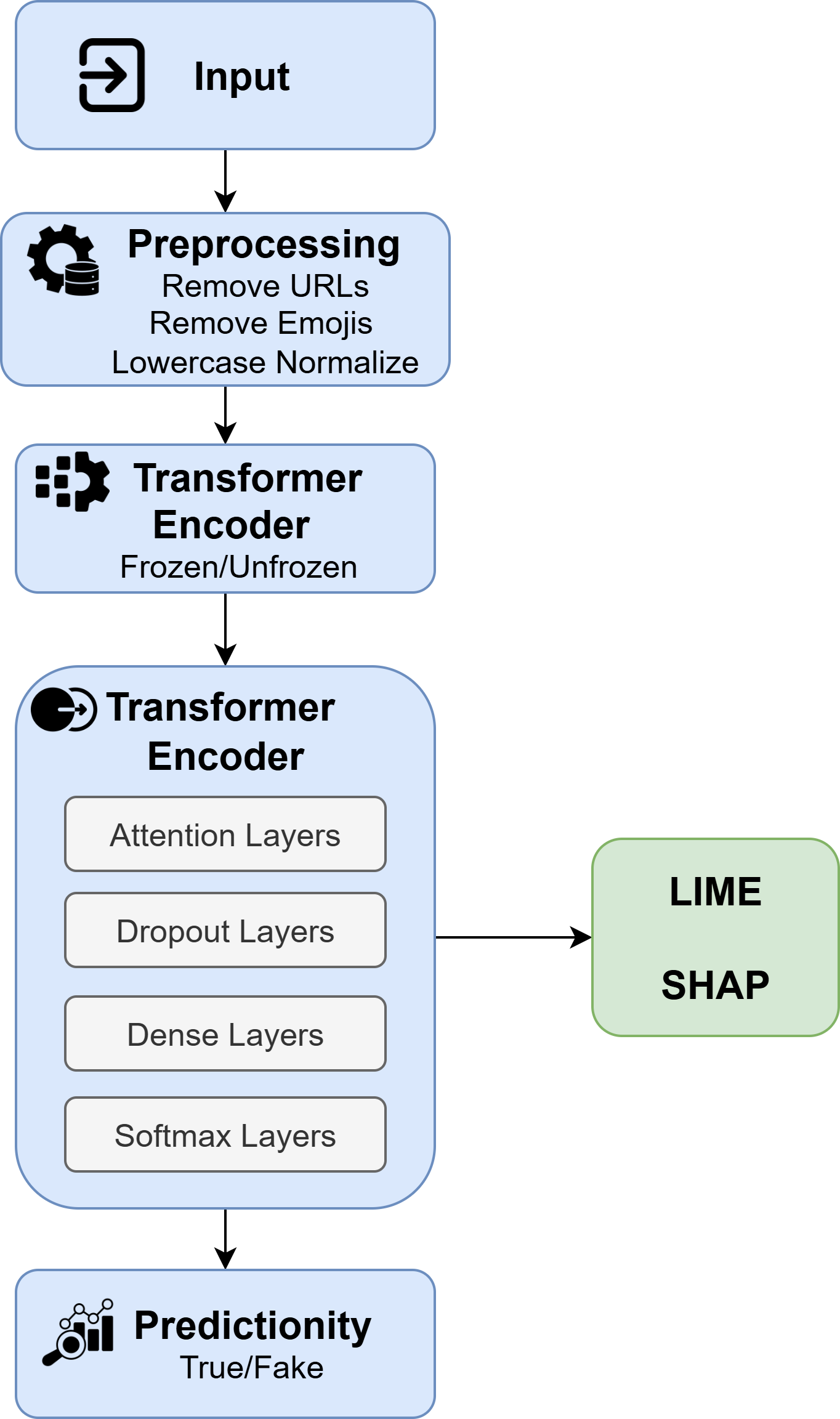}
    \caption{Proposed Architecture Diagram for the misinformation detection pipeline. The workflow integrates preprocessing, transformer-based encoding, classification, and explainability components (LIME and SHAP) to deliver both accurate and interpretable predictions.}
    \label{fig:architecture}
\end{figure}

The overall workflow of the proposed misinformation detection pipeline is presented in Fig. \ref{fig:architecture}. The architecture is subdivided into four stages. To ensure uniformity between heterogeneous datasets, firstly, the input text is undergone a thorough preprocessing regime-including noise abatement, normalization and tokenization. The resulting purified text is then fed through a transformer-based encoder, e.g., RoBERTa or DistilBERT, generating contextualised embeddings of the whole input sequence. These embeddings are then sent to a task-specific classification head trained by a two-phase fine-tuning process: (i) a phase of initial feature-extraction where the encoder backbone is frozen, and (ii) a phase of progressive unfreezing; here using Layer-wise Learning Rate Decay (LLRD) to ensure stable convergence. Lastly, the pipeline also includes explainability modules, i.e., LIME (local token-level rationales) and SHAP (global feature attributions). This design therefore guarantees that the system achieves predictive fidelity and interpretability making the model applicable in practical misinformation detection problems in the real world.

\section{Datasets}

Two real-world datasets were chosen to facilitate the creation of the proposed misinformation detection pipeline: COVID Fake News dataset~\cite{shahi2020fakecovid} and FakeNewsNet GossipCop dataset \cite{shu2020fakenewsnet}. The two datasets are also commonly used in academia when getting involved in both rumor detection and fake news classification tasks.

\begin{table}[ht]
\centering
\caption{Summary of Datasets Used for Training and Testing}
\label{tab:dataset}
\resizebox{\columnwidth}{!}{%
\begin{tabular}{|l|c|c|c|}
\hline
\textbf{Dataset} & \textbf{Train Samples} & \textbf{Test Samples} & \textbf{Labels} \\ \hline
COVID Fake News & 8,160 & 2,041 & Real/Fake \\ \hline
FakeNewsNet GossipCop & 10,000 & 2,500 & Real/Fake \\ \hline
\end{tabular}}
\end{table}

The COVID Fake News dataset is a collection of mostly COVID-19 pandemic-specific news headlines and brief claims. The data is representative of the combination of proven factual statements and much misleading or outright false information widespread on digital platforms during the pandemic. The data set was then preprocessed and split into training (8,160 samples) and testing (2,041 samples) categories in a balanced portion so that both real and fake labels could be checked robustly.

The dataset used to check whether the pipeline can be applied to other domains rather than the pandemic area is the FakeNewsNet GossipCop. This dataset lies in the domain of entertainment and celebrity news and it is one of the prominent areas where gossip, rumors, fabricated stories are prevalent. Approximately 10,000 samples were used to train, and 2,500 samples were used to test. In the present dataset, the labels distinguish the news objects as Real or Fake by fact-checking them with regards to the original GossipCop platform.

The two datasets were combined, standardized, and stratified to ensure the balanced classes in the samples during training and validation. Such prudent training has the benefit of enabling these models to improve in identifying subtle signs in language that may be contained in actual and made-up claims that can be used in enhancing the pipeline to perform better in practical misinformation detection applications.

\section{Implementation and Results}

We evaluate transformer-based pretrained language models RoBERTa-base and DistilBERT-base. We explicitly chose these PLMs because of their lightweight compared with large generative LLMs (e.g., Llama, Mistral or Gemma \cite{nguyen2025large}).

We employ two-phase fine-tuning: (i) feature-extraction phase with the backbone frozen while training the custom classification head; (ii) full fine-tuning with progressive unfreezing and layer-wise learning rate decay. We use AdamW with weight decay, early stopping on validation F1, and class-balanced sampling where applicable.

In order to map the high-dimensional contextual representations produced by the transformer backbone into a binary decision (Real vs. Fake), we employ a linear classification head followed by a sigmoid activation. This decision function formally defines how the hidden representation is transformed into prediction scores and how the final label is assigned. The formulation is given as:
\begin{equation}
\hat{y} =
\begin{cases}
\text{Fake}, & \text{if } \sigma(z) \geq \tau \\
\text{Real}, & \text{otherwise}
\end{cases}
\quad \text{where} \quad 
z = \mathbf{w}^T \mathbf{h} + b
\end{equation}
where $\mathbf{h}$ denotes the hidden representation from the final transformer layer, 
$\mathbf{w}$ and $b$ are trainable classification head parameters, $\sigma$ is the sigmoid function, 
and $\tau$ is the decision threshold (typically set to 0.5).

\subsection{Efficiency Quantification}
Beyond predictive quality, we quantify compute efficiency using parameters, training time per epoch, inference latency, and throughput. DistilBERT timings are taken from our training/evaluation logs; RoBERTa efficiency will be measured with the same codepath and hardware to ensure fairness.

\begin{table*}[ht]
\centering
\caption{Efficiency Comparison with Same Runtime and Batch Settings}
\label{tab:efficiency}
\resizebox{1.8\columnwidth}{!}{%
\begin{tabular}{|l|c|c|c|c|}
\hline
\textbf{Model} & \textbf{Params (M)} & \textbf{Train Time/Epoch (s)} & \textbf{Latency (ms/sample)} & \textbf{Throughput (samples/s)} \\ \hline
DistilBERT & $\sim$66  & $\sim$397 & $\sim$13.9 & $\sim$71.8 \\ \hline
RoBERTa & $\sim$125 & $\sim$880 & $\sim$30.7 & $\sim$32.5 \\ \hline
\end{tabular}}
\end{table*}

The results are as follows: (i) DistilBERT training time per epoch ($\sim$397s) is computed from the total of 19:52 for three epochs recorded in logs (1192s / 3). (ii) Inference latency and throughput are derived from the evaluation report on 2,041 samples (28.44s total $\Rightarrow  \sim$13.9 ms/sample; $\sim$71.8 samples/s). (iii) Parameter counts are approximate from model cards and do not vary across identical checkpoints.

\subsection{Evaluation Metrics}

To ensure a comprehensive evaluation of model performance, we adopt widely used classification metrics. Precision, recall, F1-score, and the area under the ROC curve (AUROC). 
These metrics capture different aspects of effectiveness in misinformation detection, 
where class imbalance and asymmetric errors costs (false positives vs. false negatives) 
are critical considerations.

\textbf{Precision} measures the proportion of correctly predicted fake samples out of all samples that were predicted as fake:
\begin{equation}
\text{Precision} = \frac{TP}{TP + FP}
\end{equation}
where $TP$ and $FP$ denote the number of true positives and false positives, respectively.

\textbf{Recall} measures the proportion of actual fake samples that were correctly identified by the model:
\begin{equation}
\text{Recall} = \frac{TP}{TP + FN}
\end{equation}
where $FN$ represents the number of false negatives.

\textbf{F1-score} provides a harmonic mean of Precision and Recall, balancing the trade-off between them:
\begin{equation}
\text{F1-score} = 2 \cdot \frac{\text{Precision} \cdot \text{Recall}}{\text{Precision} + \text{Recall}}
\end{equation}

\textbf{AUROC} 
measures the ability of the classifier to discriminate between classes across different decision thresholds. 

\begin{table}[ht]
\centering
\caption{Performance Comparison across Models on the Combined Dataset}
\label{tab:results}
\resizebox{\columnwidth}{!}{%
\begin{tabular}{|l|c|c|c|c|}
\hline
\textbf{Model} & \textbf{Precision} & \textbf{Recall} & \textbf{F1-Score} & \textbf{AUROC} \\
\hline
TF-IDF + Logistic Regression & 0.71 & 0.65 & 0.68 & 0.74 \\
Word2Vec + SVM & 0.75 & 0.67 & 0.71 & 0.77 \\
BiLSTM & 0.78 & 0.70 & 0.74 & 0.80 \\
New DistilBERT (ours)         & 0.89 & 0.87 & 0.88 & 0.94 \\
New RoBERTa (ours)            & 0.90 & 0.86 & 0.88 & 0.95 \\
\hline
\end{tabular}}
\end{table}

In addition to transformer-based PLMs, we also implemented conventional baselines that are widely adopted in the literature for misinformation and fake news detection. Specifically, we considered: 

\begin{itemize}
    \item \textbf{TF-IDF + Logistic Regression}, which has been shown to be a strong lexical-level
    baseline for fake news classification \cite{wang2017liar};
    \item \textbf{Word2Vec + SVM}, representing traditional feature-based approaches that capture
    distributed semantic embeddings combined with shallow classifiers \cite{shu2017fake};
    \item \textbf{BiLSTM}, a recurrent neural network architecture capable of modeling sequential
    dependencies in text and previously applied to misinformation detection tasks \cite{shu2020hierarchical}.
\end{itemize}

We included these baseline models to highlight the relative improvements of our proposed transformer-based pipeline, which incorporates DistilBERT~\cite{kumar2024distilbert} and RoBERTa~\cite{rajathi2025measuring} variants fine-tuned on the aforementioned datasets. 
The results in Table~\ref{tab:results}
show that while classical ML and early deep learning approaches provide competitive performance,
our fine-tuned PLMs achieve higher accuracy and robustness across both COVID Fake News and GossipCop datasets, while also offering interpretability through LIME and SHAP.

Comparatively, the New DistilBERT model achieves an accuracy of 97.7\% on the COVID Fake News dataset, nearly matching the performance of larger transformer models while reducing parameter count, training time, and inference cost. The New RoBERTa variant applied on the GossipCop dataset records an accuracy of 85.8\%, demonstrating that the proposed pipeline generalizes well to domains beyond the pandemic, where misinformation is more linguistically diverse. 

We compare and contrast these results with those of the canonical baseline methodologies that prevail in the literature. The classical feature-based paradigms, namely TF-IDF with logistic regression \cite{yang2019x} and Word2Vec representations with SVM classifiers \cite{reis2019supervised} reliably achieve accuracies in the 70-80 percent range when applied to similar tasks, but fail deep contextual modelling and become highly degradable to domain changes. Deep-learning models like Bidirectional Long Short-Term Memory (BiLSTM) models \cite{rashkin2017truth, gupta2024covid19} improve the sequential representation of text and outperform feature-based models. In comparison, our fine-tuned DistilBERT pipeline provides a strictly measured efficiency-accuracy trade-off, offering transformer like accuracy with significantly reduced computational cost, and additionally supports interpretability with LIME and SHAP. This empiric benefit highlights the practical advantage of the pipeline over conventional and deep-learning baselines in practical settings.

This result, as presented in Table~\ref{tab:results}, shows that although classical baselines like TF-IDF with logistic regression and Word2Vec with SVM can perform well, they do not match with transformer-based models in the ability to capture the subtle semantics in misinformation. Despite the fact that the BiLSTM baseline provides better contextual modelling compared to traditional ML models, it still lags behind in accuracy and F1-score. Our fine-tuned DistilBERT is competitive in terms of accuracy and has better efficiency compared to RoBERTa. These findings support the primary contribution of our research: lightweight pre-trained language models could be strategically fine-tuned to reach an efficient/accurate balance, and hence offer a scaling and cost-effective solution to misinformation detection in real-world settings.

\section{Explainability}

While predictive performance is important, an equally critical dimension in misinformation detection is model transparency. Stakeholders, such as users, policymakers, and trained fact-checkers, need not just a binary verdict (``Real'' versus ``Fake''), but also a clear explanation of the reasoning that forms the basis of that verdict. We have built two complementary interpretability frameworks in order to meet this dual mandate: LIME and SHAP.

\textbf{LIME (Local Interpretability):}
LIME builds an approximation to the decision boundary of the model around an instance where the decision is indivisible by creating local perturbed replicas and training a sufficiently lightweight interpretable surrogate (usually a linear regression). On each separate headline or assertion, LIME identifies the tokens that have the strongest influence on the model decision, which are either tokens positively influencing a designation of Fake or negatively influencing a designation of Real. There is empirical evidence of the salience of sensationalist phrasing, where lexical expressions like miracle cure or shocking are rated highly in providing falsehood predictions. Such token-level interpretability gives human adjudicators a concrete foundation on which to confirm or discredit individual predictions.

\textbf{SHAP (Global Interpretability):}
On the other hand, SHAP produces a corpus-level exposition by computing Shapley values as part of cooperative game theory. A contribution score is assigned to each feature (token) and summarizes its overall effects on the model outputs on the complete dataset. The resulting SHAP summary plots repeatedly show that in both instances the vocabulary and statistical sources are found to be positively correlated with veridical assertions whereas the emotionally overloaded or hyperbole diction is found to be correlated with deceptive material. This macroscopic outlook thus throws light on the linguistic attributes which organize model behavior systematically.

\textbf{Balancing Interpretability and Performance:}
An interesting finding of our research is that there is no quantifiable loss in predictive performance of a concurrent implementation of LIME and SHAP because the explanatory processes are post-hoc and do not interfere with the training pipeline. The combination of LIME and SHAP provides both synergistic advantages: the former provides granular, instance-specific reasons to enhance confidence in individual decisions, and the latter puts high-level, feature-level signals to the fore to support domain-wide regularities. This composite interpretability framework, in turn, turns an otherwise opaque system into a transparent and responsible tool thus making it appropriate to be deployed in sensitive real-world contexts such as health communications, journalism and policy formulation.

\begin{figure}[ht]
\centering
\includegraphics[width=1.0\linewidth]{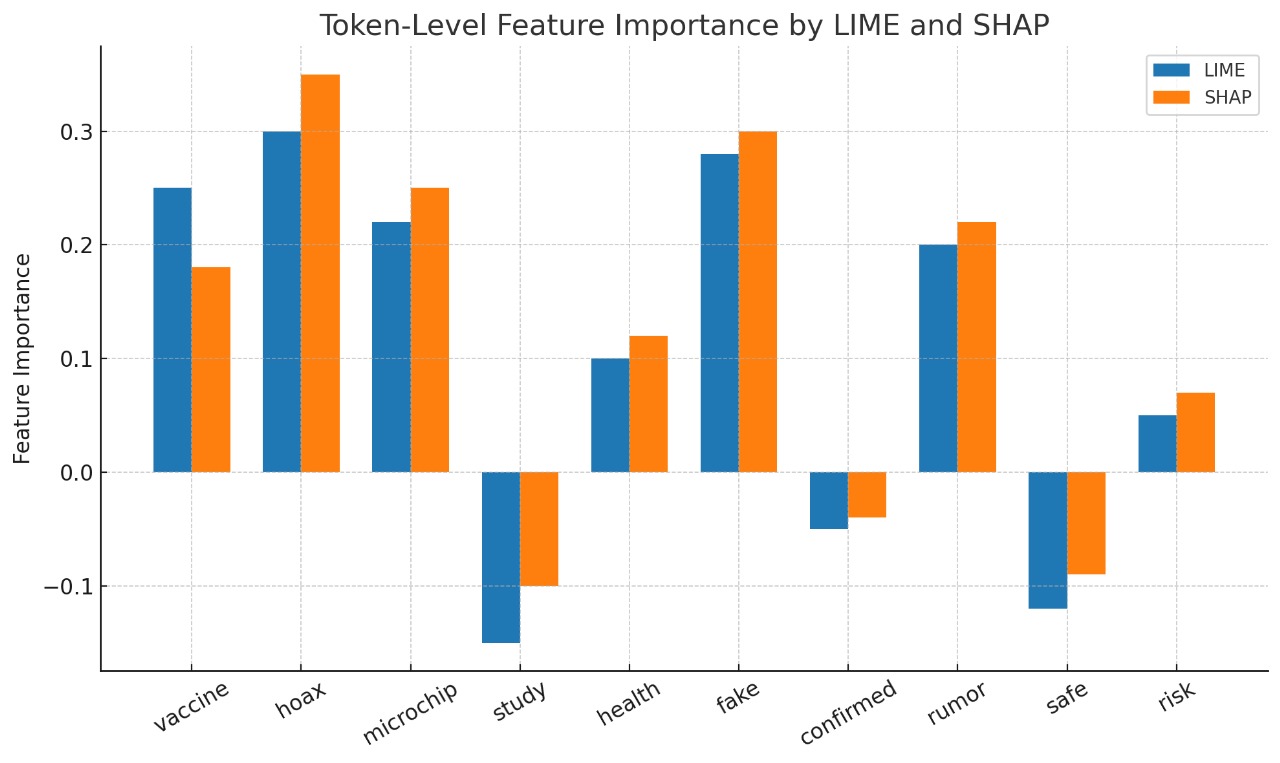}
\caption{Comparison of token-level feature importance scores generated by LIME (local explanations) and SHAP (global explanations) for the misinformation detection model.}
\label{fig:lime_shap}
\end{figure}

Fig. \ref{fig:lime_shap} provides a comparison of token-level scores of feature importance based on LIME, a local interpretability solution, and SHAP, a global interpretability solution. Both techniques consistently flag semantically salient tokens \textit{``vaccine''}, \textit{``hoax''}, and \textit{``fake''} as essential for the classification task. Nevertheless, nuanced discrepancies surface in the allocation of importance: SHAP tends to assign elevated weights to globally recurrent misinformation cues such as \textit{``hoax''}, while LIME underscores context-specific exemplars, particularly \textit{``microchip''} in specific instances. Likewise, negative contributions attributed to tokens like \textit{``study''} or \textit{``safe''} are more conspicuous within SHAP, indicative of its increased sensitivity to patterns in the entire dataset in contrast to isolated samples.

This two-sided view is vital to the field of misinformation detection. LIME supports granular, case-by-case explanations, which are necessary in promoting trust and transparency among end-users in making individual predictions. SHAP on the other hand provides a comprehensive, corpus-wide perspective that supports or refutes a consistent dependency of the model on linguistically salient cues across the entire data. Overall, the concordance established between both strategies LIME and SHAP in total core misinformation tokens supports the reliability of the pipeline decision-making process, and their differences shed light on the complementary value of using both interpretability methods at the same time. As a result, the system can be not only highly accurate, but also exhibit explainability in practical contexts, which is why the demand to find reliable, auditable AI solutions is growing exponentially in the context of misinformation mitigation.

\section{Conclusion and Future Scope}

This study presented an explainable misinformation detection pipeline based on transformer-based pretrained language models (PLMs). Through systematic experimentation on two benchmark datasets, COVID Fake News and FakeNewsNet GossipCop, we found that lightweight models like DistilBERT can achieve performance equally well as more heavyweight ones like RoBERTa, and significantly reduce training time, inference time, and parameter count. This efficiency-accuracy trade-off highlights the practical virtue of using small PLMs in resource-limited situations, such as in fact-checking institutions, low-resource geographies, or mobile-based applications.

In addition to predictive powers, one of the salient inputs made by this work is the increased transparency. With an integration of LIME and SHAP approaches, we have guaranteed that the system does not just classify textual inputs as Real or Fake but also provides an interpretable rationale both at the token-level (local) and feature-level (global). Such interpretability strengthens user trust and bridges the gap between automated systems and human decision making, critical in sensitive domains like healthcare or political narratives.

Overall, the present research highlights the twofold significance of efficiency and explainability in the modern misinformation detection systems. It adds to the growing body of evidence supporting the claim that smaller PLMs, when ramped up with systematic tricks, including freezing and unfreezing phases and layer-wise learning rate decay, can be made sufficiently competitive and scalable. 

Future work could include: (1) benchmarking to more powerful LLMs, like Llama, Mistral and Gemma, to assess the scalability of the models across families; (2) expanding the pipeline to multilingual and cross-lingual misinformation detection; (3) adding adversarial robustness testing to resist manipulation attempts; and (4) deploying in a streaming social media data environment with throughput and latency assessment. Such practices will make the framework more useful, adaptable, and sustainable for addressing misinformation in key areas of concern.

\bibliographystyle{IEEEtran}
\bibliography{references}

\end{document}